%% file: grbm.tex
\documentclass{article}

\usepackage[nonatbib,preprint]{nips_2018}

\usepackage[utf8]{inputenc} 
\usepackage[T1]{fontenc}    
\usepackage{hyperref}       
\usepackage{url}            
\usepackage{booktabs}       
\usepackage{amsfonts}       
\usepackage{nicefrac}       
\usepackage{microtype}      
\usepackage{amsmath}%
\usepackage{amssymb}%
\usepackage{graphicx}
\usepackage{subcaption}
\usepackage{algorithm}
\usepackage{algorithmic}
\usepackage{multicol}
\usepackage{cite}
\usepackage{tabularx}
\usepackage{enumitem}

\usepackage{tikz}
\usepackage{pgfplots}
\usetikzlibrary{shapes}
\usetikzlibrary{pgfplots.groupplots}
\usetikzlibrary{math}
\pgfplotsset{every tick label/.append style={font=\small}}
\pgfplotsset{compat=1.13}

\newcommand{\statements}{\mathcal{F}}

\newcommand{\sources}{\mathcal{S}}

\newcommand{\tpr}{\mathrm{tpr}}
\newcommand{\fpr}{\mathrm{fpr}}

\newcommand{\etal}{et~al.}

\title{Combining Restricted Boltzmann Machines with Neural Networks for Latent Truth Discovery}

\author{
  Klaus Broelemann\\
  SCHUFA Holding AG\\
  Wiesbaden, Germany \\
  \texttt{klaus.broelemann@schufa.de} \\
	\And
	Gjergji Kasneci\\
  SCHUFA Holding AG\\
  Wiesbaden, Germany \\
  \texttt{gjergji.kasneci@schufa.de} \\  
}

\begin{document}

\maketitle

\begin{abstract}
  Latent truth discovery, LTD for short, refers to the problem of aggregating multiple claims from various sources in order to estimate the plausibility of statements about entities.
  In the absence of a ground truth, this problem is highly challenging, when some sources provide conflicting claims and others no claims at all.
  In this work we provide an unsupervised stochastic inference procedure on top of a model that combines restricted Boltzmann machines with feed-forward neural networks to accurately infer the reliability of sources as well as the plausibility of statements about entities. In comparison to prior work our approach stands out (1) by allowing the incorporation of arbitrary features about sources and claims, (2) by generalizing from reliability per source towards a reliability function, and thus (3) enabling the estimation of source reliability even for sources that have provided no or very few claims, (4) by building on efficient and scalable stochastic inference algorithms, and (5) by outperforming the state-of-the-art by a considerable margin. 
 
\end{abstract}

\section{Introduction}

For a large number of modern applications, such as online booking, social editing, product recommendation, etc., information about the same entities is aggregated from different sources. In the absence of a ground truth, one of the main challenges when aggregating information from multiple sources is the consolidation of contradicting claims. Here, contradictions may arise from noisy, outdated, erroneous, or incomplete information; e.g., for the same flight, different flight booking web sites may report differing arrival times. 
The problem of aggregating multiple claims from various sources to estimate the plausibility of statements about entities is referred to as the \emph{latent truth discovery} (LTD). 

\paragraph{Related work}
State-of-the-art approaches to the LTD problem go well beyond computing a simple majority vote among all sources. 
Often they follow a Bayesian approach by making certain assumptions about the generative process of the observed data and jointly inferring source reliabilities and plausibility of statements from the observed claims~\cite{dong2009integrating,li2012truth,zhao2012bayesian,kasneci2011cobayes,kasneci2010bayesian,zhao2012probabilistic,bachrach2012grade}. Further approaches apply iterative fix-point methods~\cite{yin2008truth,galland2010corroborating,wang2012truth,wang2013recursive} to solve the LTD problem.
Another approach~\cite{Broelemann2017,Broelemann2018} uses restricted Boltzmann machines (RBMs) to model the latent truth discovery problem. Optimizing the likelihood of the observations is done by training RBMs. The authors show that source reliabilities and statement plausibilities are easily computed from the trained RBM.
We refer to Li~\etal~\cite{li2015survey} for a more comprehensive survey work on latent truth discovery.

The classical LTD problem can be extended by integrating additional data that is available in some settings. Several publications deal with different types of additional data, such as temporal information~\cite{Zhang2017,Yao2018}, spatio-temporal data~\cite{Garcia-Ulloa2017}, domains~\cite{Lin2018}, uncertainty of claims~\cite{wang2014using, Huang2017}, or difficulty to answer~\cite{Aydin2017}.

LTD methods typically determine the reliability of each source from the available data.
This causes a long-tail problem, when a notable number of sources have only few claims. The reliability of each source can then only be insufficiently determined. Li~\etal~\cite{li2014confidence} overcome this problem in the case of continuous data by computing confidence intervals for all sources. 

\paragraph{Our contribution}
In this work, we propose a latent truth discovery method that allows to incorporate arbitrary features. Doing so allows us to generalize from reliabilities of sources towards a reliability function that dynamically computes the reliability based on a feature vector. The reliability function allows to determine the reliability of similar sources in an combined way and thus addresses the long-tail problem.

Our work is based on former work on latent truth discovery via restricted Boltzmann machines~\cite{Broelemann2017,Broelemann2018} and extends it to incorporate a effective reliability functions, such as feed-forward neural networks.

In comparison to prior work that addresses the LTD problem our approach stands out (1) by allowing the incorporation of arbitrary features about sources and claims with a single method instead of using different methods for different sets of features, (2) by generalizing from reliability per source towards a reliability function, (3) by addressing the long-tail problem, (4) by building on efficient and scalable stochastic inference algorithms, and (5) by outperforming the state-of-the-art by a considerable margin.

Our approach can be applied to a wide range of different LTD settings with different types of features. Although this work focuses on the algorithm and its theoretical foundations, we also perform experiments on multiple datasets with different types of features. Nevertheless, a comprehensive evaluation on numerous different datasets, including a comparison with all the different extensions of the basic LTD problem for these datasets, is out of scope of this work.

The remainder of this paper is organized as follows: In Section~\ref{sec:ltd_rbm} we review the LTD approach based on restricted Boltzmann machines. Section~\ref{sec:gen_ltd} is devoted to our novel, more general approach, which combines RBMs with feed-forward neural networks to address the LTD problem. Finally, in Section~\ref{sec:experiments}, we show the evaluation results of the proposed method in comparison with state-of-the-art techniques on multiple open source datasets.

\section{Latent truth discovery model}
\label{sec:ltd_rbm}
The starting point of our approach is the LTD-RBM algorithm~\cite{Broelemann2017,Broelemann2018} that uses restricted Boltzmann machines (RBMs) to solve the latent truth discovery (LTD) problem. In this section, we review the LTD-RBM approach and introduce important terms and formulations of the basic LTD problem in \ref{subsec:ltd-formulation}. Subsequently, we briefly review restricted Boltzmann Machines in \ref{subsec:rbm}. In \ref{subsec:ltd-rbm}, we then describe the LTD-RBM algorithm that serves as foundation for the following sections.

\subsection{Truth discovery formulation}
\label{subsec:ltd-formulation}
\begin{figure}
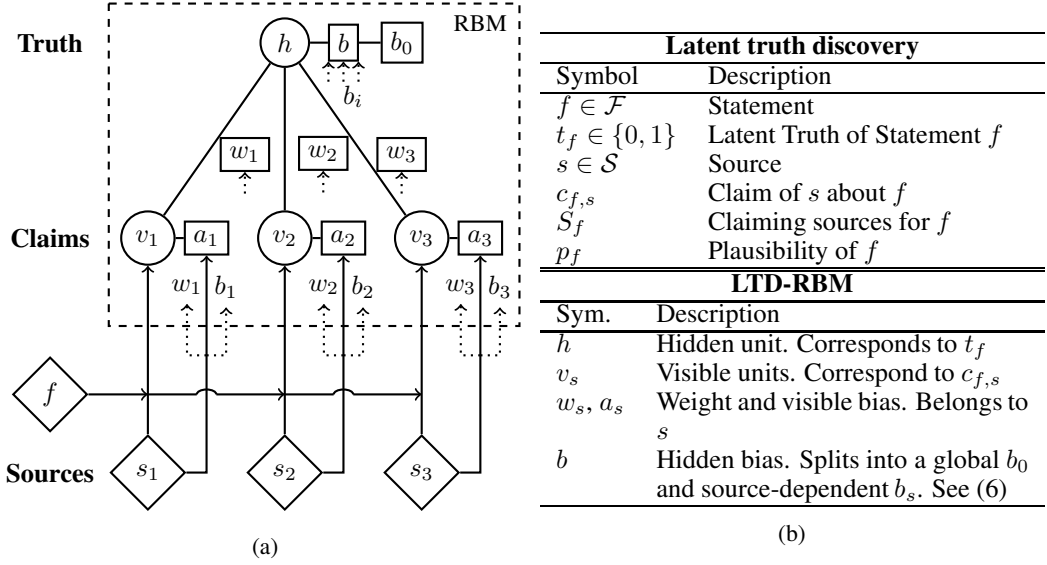
%
\begin{subfigure}[t]{0.51\textwidth}
\begin{minipage}{\textwidth}
	\include{ltd_rbm}
\end{minipage}
\vspace{-3mm}
\caption{}
\label{fig:ltd-rbm}
\end{subfigure}
\begin{subfigure}[t]{0.48\textwidth}
\begin{minipage}{\textwidth}
\begin{tabularx}{\textwidth}{lX}
	\hline
	\multicolumn{2}{c}{\bf Latent truth discovery}\\
	\hline
	Symbol & Description\\
	\hline
	$f\in\statements$   & Statement\\
	$t_f\in\{0,1\}$     & Latent Truth of Statement $f$\\
	$s\in\sources$      & Source\\
	$c_{f,s}$ & Claim of $s$ about $f$\\
	$S_f$               & Claiming sources for $f$\\
	$p_f$								& Plausibility of $f$\\
	\hline
\end{tabularx}\\
\begin{tabularx}{\textwidth}{lX}	
	\hline
	\multicolumn{2}{c}{\bf LTD-RBM}\\
	\hline
	Sym. & Description\\
	\hline
	$h$ & Hidden unit. Corresponds to $t_f$\\
	$v_s$ & Visible units. Correspond to $c_{f,s}$\\
	$w_s$, $a_s$ & Weight and visible bias. Belongs to $s$\\
	$b$ & Hidden bias. Splits into a global $b_0$ and source-dependent $b_s$. See (\ref{eqn:hidden_bias_from_sources})\\
	\hline
\end{tabularx}
\end{minipage}
\caption{}
\label{fig:list-of-symbols-1}
\end{subfigure}
\caption{(a) Schematic display of the LTM-RBM for three claims. Claims depend on the statement $f$ and on the corresponding source. Parameters depend only on the sources. {\bf Note:} Some lines are only indicated by a {\bf dotted line}, e.g. $s_1$ connects to $w_1$ and to $b$. (b) List of Symbols}%
\end{figure}

The main challenges of a truth discovery approach is to deal with conflicting data provided by different sources. For this task, the following terms are used (see also Fig.~\ref{fig:list-of-symbols-1}):

Let $\statements$  be a set of binary \textbf{statements}, i.e. statements that can be true of false. 
For each statement $f\in\statements$, we are interested in the \textbf{latent truth} $t_f\in\{0,1\}$, i.e., whether a statement is false or true.

The latent truth cannot be observed directly and knowledge about it is provided by a set $\sources$ of \textbf{sources}. For each statement $f$, we receive from a source $s$ at most one (potentially wrong) \textbf{claim} $c_{f,s}\in\{0,1\}$ about the latent truth $t_f$. Note that sources are not required to provide claims about all statements. We denote the set of actually \textbf{claiming sources} for a statement $f$ by $S_f$.

Based on the available claims, one goal of truth discovery is to compute the \textbf{plausibility} of a statement to be true. This is given by the probability
\begin{equation}
	\label{eqn:ltd:plausibility}
		p_f := P(t_f = 1|(c_{f,s})_{s\in S_f}) 
\end{equation}
\subsection{Restricted Boltzmann machines}
\label{subsec:rbm}
Restricted Boltzmann machines (RBMs) are a powerful tool for inferring hidden factors in a sequence of i.i.d. observations. In the following, we will briefly review their key features that are relevant for this work. We refer to classical literature~\cite{hinton2012rbms,hinton2002training} for a comprehensive introduction to RBMs.

An RBM is a graphical model with one visible and one hidden layer of binary units\footnote{Other types of units are also possible, but for this work, we only consider binary units.}. These layers form an undirected bipartite graph in which the edge between a visible unit $i$ and a hidden unit $j$ is attributed with a weight $w_{ij}$. Furthermore, each visible unit has a bias $a_i$ and each hidden unit has a bias $b_j$.
Given a state with values $v_i\in\{0,1\}$ and $h_i\in\{0,1\}$ for the visible units and the hidden units respectively, one can compute the following conditional probabilities: 
\begin{align}
	P(h_j=1|\mathbf{v;a,w,b}) &= \sigma\left(b_j+\sum\nolimits_iw_{ij}v_i\right) & 
	P(v_i=1|\mathbf{h;a,w,b}) &= \sigma\left(a_i+\sum\nolimits_jw_{ij}h_j\right)
	\label{eqn:rbm:probability}
\end{align}
with the logistic function $\sigma(x) = (1 + e^{-x})^{-1}$.
Given a sequence $(\mathbf{v}^k)_{k=1,\ldots,N}$ of observations for the visible layer, the weights and biases can be trained in a way that maximizes the log likelihood
\begin{equation}
	\label{eqn:rbm:likelihood}
	\mathcal{L}(\mathbf{a,w,b}) = \sum\nolimits_{k=1}^N\mathcal{L}_k(\mathbf{a,w,b}) = \sum\nolimits_{k=1}^N\log P(\mathbf{v}^k|\mathbf{a,w,b})
\end{equation}
One way to optimize the log-likelihood is to use stochastic gradient descent. This raises the problem of computing the derivative $\frac{\partial \mathcal{L}_k}{\partial w_{i,j}}$, and analogously for $a_i$ and $b_j$, which is algebraically intractable. Contrastive divergence provides a rough approximation of this gradient, which has been shown to work sufficiently well for optimizing with respect to the log-likelihood~\cite{hinton2012rbms}.
Note that in this work, we explicitly see contrastive divergence as an approximation of the gradient in order to prepare for our extension in section \ref{sec:gen_ltd}. 

As a result of this optimization, the RBM learns hidden factors of the training data in the hidden layer.

\subsection{Using RBMs for latent truth discovery}
\label{subsec:ltd-rbm}
We will briefly describe how RBMs can be applied to the LTD problem and refer to~\cite{Broelemann2018} for more details. In the next section, we will substantially extend this approach to allow for the usage of additional features.

The main idea behind the application of RBMs to the truth discovery problem is to use one visible unit for each source and exactly one hidden unit for the latent truth. The learned hidden factor corresponds to the hidden truth. Thus, we will omit the index $j$ for the hidden unit and index the visible units by their associated source $s\in\sources$. A schematic overview of the model can be seen in Fig.~\ref{fig:ltd-rbm}.

By training the RBM with claims from the sources, the plausibility $p_f$ for a statement $f$ in the resulting model can be computed as
\begin{equation}
	\label{eqn:ltd-rbm:plausibility}
	p_f \stackrel{(\ref{eqn:ltd:plausibility})}= P(h = 1|(v_s=c_{f,s})_{s\in S_f}) \\
	\stackrel{(\ref{eqn:rbm:probability})}=\sigma\left(b + \sum\nolimits_{s}w_{s}c_{f,s}\right) 
\end{equation}
The true and false positive rates of a source $s$ can be computed in a similar way by
\begin{align}
	\nonumber
	\tpr_s &= P(v_s = 1|h=1)
				\stackrel{(\ref{eqn:rbm:probability})}= \sigma\left(w_s + a_s\right)\\
	\fpr_s &= P(v_s = 1|h=0)
				\stackrel{(\ref{eqn:rbm:probability})}= \sigma\left(a_s\right)
	\label{eqn:ltd-rbm:tpr_fpr}
\end{align}

\paragraph{Missing claims} The sources are not required to provide claims for all statements. Broelemann~\etal~\cite{Broelemann2018} solved this by using visible units $v_s$ only for claiming sources $s\in S_f$ to compute probabilities in (\ref{eqn:rbm:probability}) and (\ref{eqn:ltd-rbm:plausibility}). This implied the necessary adjustment to split the hidden bias into a global part $b_0$ and source-dependent parts $b_s$.

The hidden bias for a statement $f$ can then be computed as
\begin{equation}
\label{eqn:hidden_bias_from_sources}
b = b_0 + \sum\nolimits_{s\in S_f}b_s
\end{equation}
The three parameters $\theta_s = (a_s, w_s, b_s)$ for each source and the global hidden bias $b_0$ have to be learned. 
Note that in the original work, $b_s$ was expressed as a function of $\tpr_s$ and $\fpr_s$ and thus could be computed from $a_s$ and $w_s$.
While this dependency could also be maintained for the subsequent extension in section \ref{sec:gen_ltd}, we untie it for the sake of simplicity. As a consequence, $b_s$ will in the following not depend on $a_s$ and $w_s$.

In generalized model, solving the LTD problem means to learn the correct parameters $\theta_s$ as well as $b_0$. For this, we optimize the log-likelihood function (\ref{eqn:rbm:likelihood}) by using stochastic gradient descent.
Then, a single stochastic gradient descent step for a statement $f$ consists of the following steps:
\begin{enumerate}
	\item Approximate $\Delta a_s:=\frac{\partial \mathcal{L}_f}{\partial a_{s}}$, $\Delta w_s:=\frac{\partial \mathcal{L}_f}{\partial w_{s}}$ and $\Delta b:=\frac{\partial \mathcal{L}_f}{\partial b}$  ($s\in S_f$) for the log-likelihood $\mathcal{L}_f$ of the observed claims for $f$ by means of contrastive divergence.
	\item Update the weights by adding the gradient multiplied with the learning rate. 
	The gradient for $b_0$ and $b_s$ can easily be computed by
	\begin{equation}
			\label{eqn:derive_bi}
			\Delta b_s:=\frac{\partial \mathcal{L}_f}{\partial b_s} = \frac{\partial \mathcal{L}_f}{\partial b} \cdot \frac{\partial b}{\partial b_s} = \frac{\partial \mathcal{L}_f}{\partial b} \cdot 1 = \Delta b
	\end{equation}
\end{enumerate}
 
Equation (\ref{eqn:ltd-rbm:plausibility}) shows how a trained model solves the truth discovery problem. Thus, the basic RBM-based latent truth discovery algorithm first trains an RBM with stochastic gradient descent and then computes the plausibilities of statements according to (\ref{eqn:ltd-rbm:plausibility}). It is also possible to compute the reliability of sources in terms of true and false positive rate directly from the trained RBM using (\ref{eqn:ltd-rbm:tpr_fpr}).

\section{Generalized latent truth discovery}
\label{sec:gen_ltd}
In the previous section, we described the basic truth discovery problem and an RBM-based algorithm for addressing the problem. 
This algorithm learns parameters $w_s$, $a_s$ and $b_s$ for each source. 
In the following, we will generalize this approach by introducing a reliability function $g$ that dynamically computes these three values based on some input features. 
To this end, we first introduce an extension of the data model and a modification of the truth discovery problem that generalizes from source-wise reliabilities. 
Finally, we propose our extension of the RBM-based algorithm that makes use of differentiable reliability function, such as the ones represented by feed-forward neural networks.

\begin{figure}
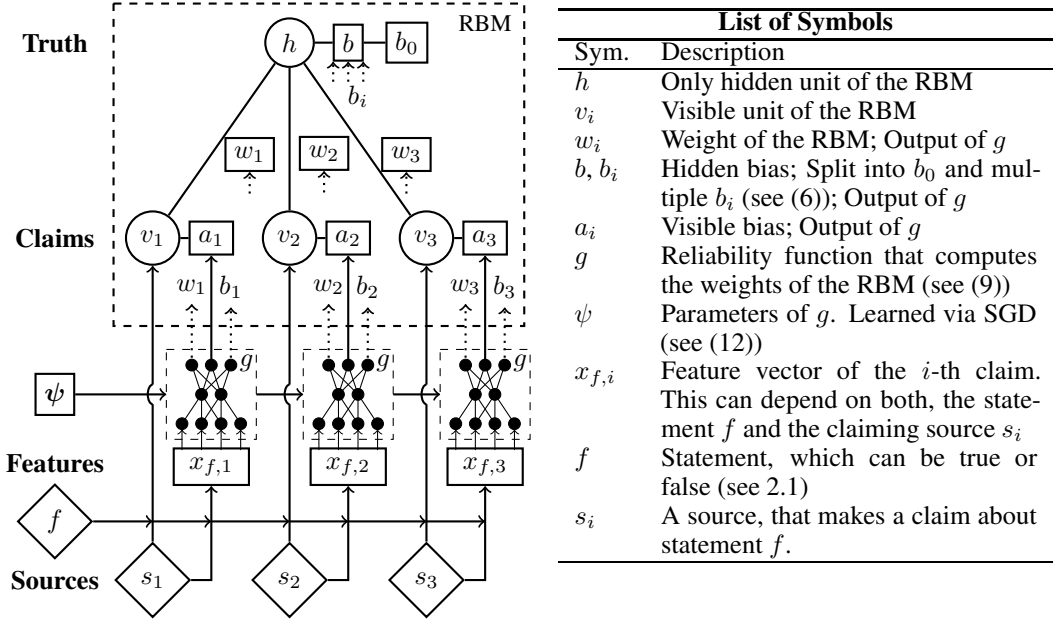
%
\begin{minipage}{0.51\textwidth}
	\include{g_ltd_rbm}
\end{minipage}
\hspace{2mm}
\begin{minipage}{0.47\textwidth}
\vspace{-10mm}
\begin{tabularx}{\textwidth}{lX}
	\hline
	\multicolumn{2}{c}{\bf List of Symbols}\\
	\hline
	Sym. & Description\\
	\hline
	$h$ & Only hidden unit of the RBM\\
$v_i$ & Visible unit of the RBM\\
$w_i$ & Weight of the RBM; Output of $g$\\
$b$, $b_i$  & Hidden bias; 
				Split into $b_0$ and multiple $b_i$ (see (\ref{eqn:hidden_bias_from_sources})); 
				Output of $g$\\
$a_i$ & Visible bias; Output of $g$\\
	$g$ & Reliability function that computes the weights of the RBM (see (\ref{eqn:grbn:function_g}))\\
	$\psi$ & Parameters of $g$. Learned via SGD (see (\ref{eqn:inner_sgd}))\\
	$x_{f,i}$ & Feature vector of the $i$-th claim. This can depend on both, the statement $f$ and the claiming source $s_i$\\
	$f$ & Statement, which can be true or false (see \ref{subsec:ltd-formulation})\\
	$s_i$ & A source, that makes a claim about statement $f$.\\
	\hline
\end{tabularx}
\end{minipage}
\vspace{-5mm}
\caption{Schematic display of the generalized LTM-RBM for three claims. Claims depend on the statement $f$ and on the corresponding source $s_i$.
{\bf Note:} Some lines are only indicated by a {\bf dotted line}, e.g. $g$ connects to $w_i$ and to $b$.}
\label{fig:g-ltd-rbm}%
\vspace{-5mm}
\end{figure}

\subsection{Extended model}
\label{subsec:gltd-formulation}
The basic model required only sources, statements and claims as the connection between both. We extend this model by adding feature vectors $\mathbf{x}_{f,s}$ for each statement $f$ and claiming source $s\in S_f$. The features can depend on sources, on statements or on the combination of both.
Typical examples for features are: the field of expertise of a source, the topic of a statement, the difficulty of a statement, and the certainty of a source about its claim.

Using such feature vectors leads to a modified truth discovery problem. Instead of trying to compute the plausibilities based on the sources and claims as described in (\ref{eqn:ltd:plausibility}), we propose the following modification:
\begin{equation}
	\label{eqn:gltd:plausibility}
	p_f := P(t_f = 1|(c_{f,i})_{i=1,\ldots,n_f},(\mathbf{x}_{f,i})_{i=1,\ldots,n_f})
\end{equation}
This formula contains two significant changes: 
\begin{itemize}[leftmargin=*,topsep=0pt]
	\item The plausibilities are estimated with respect to the feature vectors. This allows to adjust source reliabilities for each statement, e.g. depending on the statements topic.
	\item Claims and feature vectors are not indexed by the source, but by a running number. This means that the sources of claims do not need to be identified, as long as the corresponding feature vector is known. This addresses the long-tail problem. Plausibilities can be computed, even if sources have been unknown before (i.e., the typical cold-start problem). In the following, we use the running index $i$ for the generalized model and refer to the corresponding source as $s_i$.
\end{itemize}

\subsection{RBMs and neural networks for generalized truth discovery}
In subsection \ref{subsec:ltd-rbm} we described an algorithm for truth discovery based on RBMs. 
In this model, each source $s\in\sources$ was represented by three parameters $\theta_s = (a_s, w_s, b_s)$. In the following, we will generalize this model. More specifically,
instead of storing one parameter vector for each source, we represent the parameters by a reliability function:
\begin{equation}
\label{eqn:grbn:function_g}
\theta_i^f = g(\mathbf{x}_{f,i};\boldsymbol{\psi}) = \left(\begin{array}{l}g_a(\mathbf{x}_{f,i};\boldsymbol{\psi})\\g_w(\mathbf{x}_{f,i};\boldsymbol{\psi})\\g_b(\mathbf{x}_{f,i};\boldsymbol{\psi})\end{array}\right)
\end{equation}
where $\boldsymbol{\psi}$ are the parameters of the reliability function. 
Using $g$ allows to base the reliabilities on feature vectors $\mathbf{x}_{f,i}$ instead of specific sources. Note, that the weight-vector $\theta$ now depends on both $s_i$ and $f$. That means, that the reliability of a source depends on the statement. One could for example think of lower reliabilities for hard-to-decide problems or topic-dependent reliabilities of the sources. We show a schematic overview of the proposed method in Fig.~\ref{fig:g-ltd-rbm}.

Using definition (\ref{eqn:grbn:function_g}), the generalized method allows us to compute the plausibilities introduced in (\ref{eqn:gltd:plausibility}), as we will show in the following.
Given a statement $f$, the claims $c_{f,i}$ ($i=1,\ldots,n_f$) and the corresponding feature vectors, we can compute the plausibility $p_f$ as follows:
\begin{align}
	p_f & \stackrel{(\ref{eqn:gltd:plausibility})}= P(t_f = 1|(c_{f,i})_{i=1,\ldots,n_f},(\mathbf{x}_{f,i})_{i=1,\ldots,n_f}) \nonumber\\
			& \stackrel{(\ref{eqn:rbm:probability})}= \sigma\left(b + \sum_{i=1}^{n_f}c_{f,i}w_i\right) 
			\stackrel{(\ref{eqn:grbn:function_g})}= \sigma\left(b_0 + \sum_{i=1}^{n_f} g_b(\mathbf{x}_{f,i};\psi) + \sum_{i=1}^{n_f}c_{f,i}g_w(\mathbf{x}_{f,i})\right) \label{eqn:grbm:plausibility}
\end{align}

This reduces LTD to finding suitable values for $b_0$ and $\boldsymbol{\psi}$. Like in subsection \ref{subsec:ltd-rbm}, we achieve this by optimizing the log-likelihood function
\begin{equation}
	\label{eqn:objective_function2}
	\mathcal{L} = \sum\nolimits_{f\in\statements}\mathcal{L}_f = \sum\nolimits_{f\in\statements}\log P((c_{f,i})|\boldsymbol{\psi},b_0)
\end{equation}
Again, we use stochastic gradient descent. For this, we need to compute the derivatives
\begin{align}
	& \Delta \psi_j := \frac{\partial}{\partial \psi_j}\mathcal{L} 
	= \left(\frac{\partial}{\partial \theta_1^f}\mathcal{L}_f,\ldots,\frac{\partial}{\partial \theta_m^f}\mathcal{L}_f\right)\cdot\left(\frac{\partial \theta_1^f}{\partial \psi_j},\ldots,\frac{\partial \theta_m^f}{\partial \psi_j}\right)^t 
	\nonumber\\
	\stackrel{(\ref{eqn:grbn:function_g})}{=}{}& \sum_{i=1,\ldots,n_f}\Delta a_i \cdot \frac{\partial }{\partial \psi_j}g_a(\mathbf{x}_{f,i};\boldsymbol{\psi}) 
		+\Delta w_i \cdot \frac{\partial }{\partial \psi_j}g_w(\mathbf{x}_{f,i};\boldsymbol{\psi}) 
		+\Delta b_i \cdot \frac{\partial }{\partial \psi_j}g_b(\mathbf{x}_{f,i};\boldsymbol{\psi}) 
		\label{eqn:inner_sgd}
\end{align}
Since $b_0$ does not depend on $\boldsymbol{\psi}$, $\Delta b_0$ is computed as shown (\ref{eqn:derive_bi}) for the basic model.

As discussed before, the derivatives $\Delta w_i$, $\Delta a_i$ and $\Delta b_i$ can be roughly estimated by contrastive divergence. Thus, given a function $g$ that can be derived with respect to the parameters $\boldsymbol{\psi}$, the log-likelihood can be optimized using stochastic gradient descent.

To summarize, we obtain the following steps for generalized truth discovery:
\begin{enumerate}[topsep=0pt]
	\item Train the reliability function $g$ (see Algorithm~\ref{alg:training})
	\item Compute the plausibilities $p_f$ using equation (\ref{eqn:grbm:plausibility}).
\end{enumerate}

\begin{algorithm}[bt]
\caption{Generalized RBM Training}\label{alg:training}
\begin{multicols}{2}
\begin{algorithmic}[1]
\FUNCTION{TrainGeneralizedRBM}{}
\STATE $\lambda \gets \text{Learning rate}$
\STATE $(c_{f,i}) \gets \text{Claims about statement }f$
\STATE $(x_{f,i}) \gets \text{Feature descriptions of the claims}$
\STATE Initialize $b_0$, PreTrain $\boldsymbol{\psi}$
\REPEAT
\FORALL{$f\in\statements$}
	\FOR{$i = 1,\ldots,n_f$}
		\STATE $a_i \gets g_a(\mathbf{x}_{f,i};\boldsymbol{\psi})$
		\STATE $w_i \gets g_w(\mathbf{x}_{f,i};\boldsymbol{\psi})$
		\STATE $b_i \gets g_b(\mathbf{x}_{f,i};\boldsymbol{\psi})$
	\ENDFOR
	\STATE $\Delta\boldsymbol{\theta} \gets$ContrastiveDivergence()
	\STATE $\Delta\boldsymbol{\psi} \gets$ SGD(error = $\Delta\boldsymbol{\theta}$)
	\FOR{$j$}
		\STATE $\psi_j \gets \psi_j + \lambda\Delta \psi_j$
	\ENDFOR
	\STATE $b_0 \gets b_0 + \lambda\Delta b_0$
\ENDFOR
\UNTIL $b_0,\boldsymbol{\psi}$ converges
\STATE {\bfseries return:} $b_0,\boldsymbol{\psi}$
\ENDFUNCTION
\vspace{1.5em}
\FUNCTION{ContrastiveDivergence}{}
	\FOR{$i = 1,\ldots,n_f$}
		\STATE $v^{(0)}_i \gets c_{f,i}$
	\ENDFOR
	\STATE Sample $h^{(0)}$ based on $P(h^{(0)} = 1|v^{(0)})$
	\STATE Sample $\boldsymbol{v}^{(1)}$ based on $P(v_i^{(1)} = 1|h^{0})$
	\STATE Sample $h^{(1)}$ based on $P(h^{(1)} = 1|v^{(1)})$
	\FOR{$i = 1,\ldots,n_f$}
		\STATE Approx. $\Delta b \approx h^{(0)} - h^{(1)}$
		\STATE Approx. $\Delta a_i \approx v_i^{(0)} - v_i^{(1)}$
		\STATE Approx. $\Delta w_i \approx v_i^{(0)}h^{(0)} - v_i^{(1)}h^{(1)}$
		\STATE $\Delta b_i \stackrel{(\ref{eqn:derive_bi})}= \Delta b$  ($i = 0,\ldots,n_f$)
	\ENDFOR
	\STATE {\bfseries return:} $\Delta\theta = (\Delta b_0,\Delta a, \Delta w, \Delta b)$
\ENDFUNCTION

\end{algorithmic}
\end{multicols}
\end{algorithm}

\paragraph{Neural networks}
The generalized RBM for truth discovery allows various classes of reliability function $g$. A sufficient requirement is being able to compute $\frac{\partial g}{\partial \boldsymbol{\psi}}$ for stochastic gradient descent. Deep feed-forward neural networks have shown to be a powerful tool for numerous applications. When using neural networks for $g$, the output layer consists of three neurons representing $g_a$, $g_w$ and $g_b$, respectively.
During training, the error terms produced by contrastive divergence are backpropagated into the neural network. Combining contrastive divergence with backpropagation in this way, we train the neural network unsupervisedly.

\paragraph{Pre-training}
Using RBMs for truth discovery can produce two dual results. In one case a true statement is encoded by $1$ in the hidden unit, in the other case, it is encoded by $0$. Both are correct results of the truth discovery: one is optimistic (sources tend to make correct claims), the other one is pessimistic (sources tend to make incorrect claims).

In order to enforce the optimistic outcome, the weights in the basic RBM model can be initialized in a way that the true positive rate is high and the false positive rate is low for all sources (or vice-versa, in case the pessimistic outcome is preferred).
From (\ref{eqn:ltd-rbm:tpr_fpr}) with (\ref{eqn:rbm:probability}), we obtain
\begin{align*}
	\tpr_s & = \sigma(a_s + w_s) = \frac{1}{1 + e^{-a_s - w_s}} & 
	\fpr_s & = \sigma(a_s) = \frac{1}{1 + e^{-a_s}}\\
	\Rightarrow a_s & = \log(\fpr_s)-\log(1-\fpr_s) & 
	w_s & = \log(\tpr_s)-\log(1-\tpr_s) - a_s
\end{align*}
In the basic RBM model~\cite{Broelemann2017,Broelemann2018}, one initializes all sources as desired using these formulas.
In order to achieve a similar behavior in the generalized model, the reliability function must be pre-trained to produce the desired initial results. The desired initial weights $\theta_i = (a_i, w_i, b_i)$ can be computed in the same way as in the basic model. Features from the training data are taken to pre-train $g$ in a supervised way to produce the desired output $(a_i, w_i, b_i)$.
One can easily modify the pre-training to incorporate domain knowledge, e.g. different weights for different classes of sources or decreasing reliabilities depending on the source-stated certainty in the truth of their claim.

In summary, our generalized approach is based on {\bfseries unsupervised training} of feed-forward networks, using contrastive divergence of the RBM ontop of pack-propagation in feed-forward networks, while the {\bfseries pre-training} is done in a {\bfseries supervised} way.

\section{Experimental evaluation}
\label{sec:experiments}
In the previous section, we introduced a generalized RBM algorithm for truth discovery as well as a learning method based on stochastic gradient descent. In the following, we compare this method with state-of-the-art algorithms and report results on its effectiveness.

\subsection{Datasets}
For this purpose, we took datasets that can be used for both the basic data model (see section \ref{subsec:ltd-formulation}) and the generalized data model (see section \ref{subsec:gltd-formulation}). The datasets are two sets of editing data from wikipedia: a population dataset and a biography datasets\footnote{see \url{https://cogcomp.cs.illinois.edu/page/resource_view/16}}. These datasets contain claims from Wikipedia users about attributes of an info box. Additionally, we also use a quiz dataset~\cite{Aydin2017} with user answers for tv quiz show questions.

We created two versions of the biography dataset to test on different characteristics. The original dataset contains a flag for each claim that indicates whether it comes from a change of the value of an attribute of interest, a change somewhere in the info box or a change somewhere else on the website. Based on this flag, we created two versions: (1) only containing changes at the attribute of interest and (2) also containing changes somewhere else in the info box. All other claims were ignored. 

Table~\ref{tab:datasets} (a) compares the characteristics of the different datasets. The long-tail problem is indicated by the significant number of 1-claim-sources, i.e. sources, that only provide one claim.

All datasets contain multinomial data, i.e. a claim provides an attribute value for an entity, e.g. a value for the population of Boston in the year 2006. We employ one-hot-encoding in order to work with binary statements, i.e. we created one statement for each entity-attribute-value combination, e.g. ``The population of Boston in 2006 was 590,763''.

\begin{table}[bt]
\centering
\caption{Evaluation setup and results.}
\label{tab:datasets}
\begin{minipage}{0.67\textwidth}
\begin{tabular}{|l|rrrr|}
\hline
 & Biogr. 1 & Biogr. 2 & Population & Quiz\\
\hline
Sources      & 46,500 & 199,254 & 4,264 & 37332 \\
Statements   & 197,734 & 197,734 & 45,468 & 1891\\
Claims       & 228,035 & 936,296  & 50,561 & 214658\\
1-Claim- & 31,336 & 110,340 & 3,224 & 12674 \\
\hspace{2mm}Sources &&&&\\
\# Features & 16 & 17 & 9 & 36\\
\hline\multicolumn{1}{c}{}\\[-2.5mm]
\multicolumn{5}{c}{(a) Dataset characteristics}\\[6.2mm]
\hline
Method & Biogr. 1 & Biogr. 2 & Population & Quiz\\
\hline
GRBM     & \bf 88.92 \% & \bf 89.78 \% &     85.00 \% & \bf 90.85 \%\\
RBM\cite{Broelemann2018}  &     81.36 \% &     87.79 \% &     80.35 \% &     84.42 \%\\
MLE\cite{wang2012truth} &     82.14 \% &     88.77 \% & \bf 85.33 \% &     29.10 \%\\
LTM\cite{zhao2012bayesian} &     82.97 \% &     88.75 \% &     70.78 \% &     88.90 \%\\
\hline\multicolumn{1}{c}{}\\[-2.5mm]
\multicolumn{5}{c}{(b) Accuracy of our method (GRBM) and reference methods.}
\end{tabular}
\end{minipage}
\begin{minipage}{0.31\textwidth}
\begin{tabularx}{\textwidth}{|X|}
\hline
{\bf Population:} \\
- Temporal order of claims\\- Registered-User-Flag\\
\hline
{\bf Biogr. 1:} \\
- \emph{Features of Population}\\
- Attribute type\\
\hline
{\bf Biogr. 2:} \\
- \emph{Features of Biogr. 1}\\
- Flag: Did the user edit the attribute?\\
\hline
{\bf Quiz:} \\
- User Confidence\\
- Question Difficulty\\
- Question Category\\
- Answering Time\\
\hline\multicolumn{1}{c}{}\\[-2.5mm]
\multicolumn{1}{c}{(c) List of Features}
\end{tabularx}
\end{minipage}
\end{table}

\paragraph{Features}
In order to apply our generalized method, we need to provide features for each claim. For this purpose, we computed general statistics, such as the number of claims from each source or the number of claims for the current statement. In addition, we included a number of dataset-specific features that can be seen in Tab.~\ref{tab:datasets} (c).

\subsection{Experiments}
In our experiments, we compared our method (GRBM) with a maximum likelihood estimation method (MLE)~\cite{wang2012truth}, the latent truth model (LTM)~\cite{zhao2012bayesian}, and the basic method based on RBMs~\cite{Broelemann2017,Broelemann2018}.
We model the reliability function for GRBM as feed-forward neural network with dataset-dependent architectures.

\paragraph{Ground truth} All our datasets come with carefully created ground-truth data for a subset of the statements. This ground-truth data is solely used for evaluation purposes -- the truth discovery is done purely unsupervised. Using the ground-truth data allows to evaluate the accuracy of the detected truth. We give the results in Tab.~\ref{tab:datasets} (b).

The experiments show that our method can benefit from a rich feature description and outperforms state-of-the-art methods. Only on one dataset, it is out-performed by MLE, but still notably better than other methods. Note that this happend for the population dataset, which carries lowest number of features. Since our dataset does not store reliability values for each source, it has to compensate the missing source information with rich features and an expressive reliability function.

\section{Conclusion}
To the best of our knowledge, this is the first work that employs stochastic gradient descent on top of RBMs and feed-forward neural networks to address the generalized LTD problem, which includes source and statement features. The superior effectiveness of the proposed method in comparison to state-of-the-art approaches results on the one hand from a more expressive description of sources and statements by means of features and on the other hand from a powerful inference based on contrastive divergence and stochastic gradient descent. We hope that the proposed approach will pave the way for more expressive and effective solutions to the LTD problem.

Based on this work, we plan to investigate the effectiveness of our approach for categorical values that change over time. Given claims with temporal information (e.g. the time when the claim was made or the time for which the claim was made), it is possible to replace the feed-forward networks by recurrent networks in order to exploit temporal dynamics.

\bibliographystyle{abbrv}
\bibliography{references}

\end{document}

%% file: ltd_rbm.tex
\begin{tikzpicture}[scale=0.26]

\tikzmath{
	\yH = 16;        
	\yV = \yH - 10;  
	\yf = \yV - 8;   
	\yS = \yf - 4;   
}

\node at (-5, \yS) {\bf Sources};
\node[thick,diamond,draw] (f) at (-5, \yf) {$f$};
\node at (-5, \yV) {\bf Claims};
\node at (-5, \yH) {\bf Truth};

\node[thick,diamond,draw, minimum size=1] (S1) at  (0, \yS) {$s_1$};
\node[thick,diamond,draw, minimum size=1] (S2) at  (7, \yS) {$s_2$};
\node[thick,diamond,draw, minimum size=1] (S3) at  (14, \yS) {$s_3$};

\node[thick,circle,draw, minimum size=1] (V1) at  (0, \yV) {$v_1$};
\node[thick,circle,draw, minimum size=1] (V2) at  (7, \yV) {$v_2$};
\node[thick,circle,draw, minimum size=1] (V3) at  (14, \yV) {$v_3$};

\node[thick,draw, minimum size=1] (a1) at  (3,  \yV) {$a_1$};
\node[thick,draw, minimum size=1] (a2) at  (10, \yV) {$a_2$};
\node[thick,draw, minimum size=1] (a3) at  (17, \yV) {$a_3$};

\node (w1_) at ($(a1) - (1,2.5)$) {$w_1$};
\node (w2_) at ($(a2) - (1,2.5)$) {$w_2$};
\node (w3_) at ($(a3) - (1,2.5)$) {$w_3$};

\node (b1_) at ($(a1) - (-1,2.5)$) {$b_1$};
\node (b2_) at ($(a2) - (-1,2.5)$) {$b_2$};
\node (b3_) at ($(a3) - (-1,2.5)$) {$b_3$};

\node[thick,circle,draw, minimum size=1] (H) at  (7, \yH) {$h$};

\node[thick,draw, minimum size=1] (b) at  (10, \yH) {$b$};

\node[minimum size=1] (b_i) at  (10.5, \yH-2.75) {$b_i$};

\node[thick,draw, minimum size=1] (b0) at  (13, \yH) {$b_0$};

\draw [thick] (V1) -- (H) node[pos=0.4, right=5, thick,draw, minimum size=1] (w1) {$w_1$};
\draw [thick] (V2) -- (H) node[pos=0.4, right=5, thick,draw, minimum size=1] (w2) {$w_2$};
\draw [thick] (V3) -- (H) node[pos=0.4, right=5, thick,draw, minimum size=1] (w3) {$w_3$};

\draw [thick] (V1) -- (a1);
\draw [thick] (V2) -- (a2);
\draw [thick] (V3) -- (a3);

\draw [thick] (H) -- (b) -- (b0);

\draw [thick, dashed] (-2,\yH + 2) -- (19,\yH + 2) -- (19, \yV - 4.5) -- (-2, \yV - 4.5) -- (-2, \yH + 2);
\node at (17,\yH + 1) {\small RBM};

\draw [thick,->] (S1) -- (V1);
\draw [thick,->] (S2) -- (V2);
\draw [thick,->] (S3) -- (V3);

\draw [thick,->] (f) -- (0,\yf);

\draw [thick] (0,\yf) -- (2.5,\yf);
\draw [thick] (3.5,\yf) arc (0:180:0.5 and 0.25);
\draw [thick,->] (3.5,\yf) -- (7,\yf);

\draw [thick] (7,\yf) -- (9.5,\yf);
\draw [thick] (10.5,\yf) arc (0:180:0.5 and 0.25);
\draw [thick,->] (10.5,\yf) -- (14,\yf);

\draw [thick,->] (S1.east) -| (a1);
\draw [thick,->] (S2.east) -| (a2);
\draw [thick,->] (S3.east) -| (a3);

\draw [thick,->,dotted] ($(a1) - (0,6)$) -| (w1_);
\draw [thick,->,dotted] ($(a1) - (0,6)$) -| (b1_);
\draw [thick,->,dotted] ($(a2) - (0,6)$) -| (w2_);
\draw [thick,->,dotted] ($(a2) - (0,6)$) -| (b2_);
\draw [thick,->,dotted] ($(a3) - (0,6)$) -| (w3_);
\draw [thick,->,dotted] ($(a3) - (0,6)$) -| (b3_);

\draw [thick,->,dotted] ($ (w1) - (0,2) $) -- (w1);
\draw [thick,->,dotted] ($ (w2) - (0,2) $) -- (w2);
\draw [thick,->,dotted] ($ (w3) - (0,2) $) -- (w3);

\draw [thick,->,dotted] ($ (b) - (0,2) $) -- (b);
\draw [thick,->,dotted,transform canvas={xshift=-2mm}] ($ (b) - (0,2) $) -- (b);
\draw [thick,->,dotted,transform canvas={xshift=2mm}] ($ (b) - (0,2) $) -- (b);

\end{tikzpicture}

%% file: g_ltd_rbm.tex
\begin{tikzpicture}[scale=0.26]
\tikzmath{
	\yH = 16;        
	\yV = \yH - 10;  
	\yg = \yV - 8.0; 
	\yX = \yg - 3.5; 
	\yf = \yX - 3;   
	\yS = \yf - 3;   
}

\tikzmath{
	\ygA = \yg - 1.5;
	\ygB = \yg;
	\ygC = \yg + 1.5;
}

\node at (-5, \yS) {\bf Sources};
\node[thick,diamond,draw] (f) at (-5, \yf) {$f$};
\node at (-5, \yX) {\bf Features};
\node[thick,draw] (psi) at (-5, \yg) {$\boldsymbol{\psi}$};
\node at (-5, \yV) {\bf Claims};
\node at (-5, \yH) {\bf Truth};

\node[thick,diamond,draw, minimum size=1] (S1) at  (0, \yS) {$s_1$};
\node[thick,diamond,draw, minimum size=1] (S2) at  (7, \yS) {$s_2$};
\node[thick,diamond,draw, minimum size=1] (S3) at  (14, \yS) {$s_3$};

\node [thick,rectangle,draw, minimum size=1, minimum width=1cm] (X1) at  (3, \yX - 0.2) {$x_{f,1}$};
\node [thick,rectangle,draw, minimum size=1, minimum width=1cm] (X2) at  (10, \yX - 0.2) {$x_{f,2}$};
\node [thick,rectangle,draw, minimum size=1, minimum width=1cm] (X3) at  (17, \yX - 0.2) {$x_{f,3}$};

\node[thick,circle,draw, minimum size=1] (V1) at  (0, \yV) {$v_1$};
\node[thick,circle,draw, minimum size=1] (V2) at  (7, \yV) {$v_2$};
\node[thick,circle,draw, minimum size=1] (V3) at  (14, \yV) {$v_3$};

\node[thick,draw, minimum size=1] (a1) at  (3,  \yV) {$a_1$};
\node[thick,draw, minimum size=1] (a2) at  (10, \yV) {$a_2$};
\node[thick,draw, minimum size=1] (a3) at  (17, \yV) {$a_3$};

\node[thick,circle,draw, minimum size=1] (H) at  (7, \yH) {$h$};

\node[thick,draw, minimum size=1] (b) at  (10, \yH) {$b$};
\node[minimum size=1] (b_i) at  (10.5, \yH-2.75) {$b_i$};

\node[thick,draw, minimum size=1] (b0) at  (13, \yH) {$b_0$};

\draw [thick] (V1) -- (H) node[pos=0.4, right=5, thick,draw, minimum size=1] (w1) {$w_{1}$};
\draw [thick] (V2) -- (H) node[pos=0.4, right=3.5, thick,draw, minimum size=1] (w2) {$w_{2}$};
\draw [thick] (V3) -- (H) node[pos=0.4, right=5, thick,draw, minimum size=1] (w3) {$w_{3}$};

\node (w1_) at ($(a1) - (1,2.5)$) {$w_{1}$};
\node (w2_) at ($(a2) - (1,2.5)$) {$w_{2}$};
\node (w3_) at ($(a3) - (1,2.5)$) {$w_{3}$};

\node (b1_) at ($(a1) - (-1,2.5)$) {$b_{1}$};
\node (b2_) at ($(a2) - (-1,2.5)$) {$b_{2}$};
\node (b3_) at ($(a3) - (-1,2.5)$) {$b_{3}$};

\draw [thick] (V1) -- (a1);
\draw [thick] (V2) -- (a2);
\draw [thick] (V3) -- (a3);

\draw [thick] (H) -- (b) -- (b0);

\draw [thick, dashed] (-2,\yH + 2) -- (19,\yH + 2) -- (19, \yV - 4.5) -- (-2, \yV - 4.5) -- (-2, \yH + 2);
\node at (17,\yH + 1) {\small RBM};

\draw [thick] (S1) -- (0, \yg - 0.5);
\draw [thick] (0, \yg + 0.5) arc (90:270:0.25 and 0.5);
\draw [thick,->] (0, \yg + 0.5) -- (V1);

\draw [thick] (S2) -- (7, \yg - 0.5);
\draw [thick] (7, \yg + 0.5) arc (90:270:0.25 and 0.5);
\draw [thick,->] (7, \yg + 0.5) -- (V2);

\draw [thick] (S3) -- (14, \yg - 0.5);
\draw [thick] (14, \yg + 0.5) arc (90:270:0.25 and 0.5);
\draw [thick,->] (14, \yg + 0.5) -- (V3);

%
%

\draw [thick,->] (psi) -- (0.7,\yg);
\draw [thick,->] (5.3,\yg) -- (7.7,\yg);
\draw [thick,->] (12.3,\yg) -- (14.7,\yg);

\draw [thick,->] (f) -- (0,\yf);
\draw [thick,->] (0,\yf) -- (3,\yf);
\draw [thick,->] (3,\yf) -- (7,\yf);
\draw [thick,->] (7,\yf) -- (10,\yf);
\draw [thick,->] (10,\yf) -- (14,\yf);
\draw [thick,->] (14,\yf) -- (17,\yf);

\draw [thick,->] (S1.east) -| (X1);
\draw [thick,->] (S2.east) -| (X2);
\draw [thick,->] (S3.east) -| (X3);

\foreach \x in {0, 7, 14} {
	\filldraw (\x + 1.5,\ygA) circle (0.3);
	\filldraw (\x + 2.5,\ygA) circle (0.3);
	\filldraw (\x + 3.5,\ygA) circle (0.3);
	\filldraw (\x + 4.5,\ygA) circle (0.3);

	\filldraw (\x + 2.5,\ygB) circle (0.3);
	\filldraw (\x + 3.5,\ygB) circle (0.3);

	\filldraw (\x + 2,\ygC) circle (0.3);
	\filldraw (\x + 3,\ygC) circle (0.3);
	\filldraw (\x + 4,\ygC) circle (0.3);

	\draw (\x + 1.5,\ygA) -- (\x + 2.5,\ygB);
	\draw (\x + 2.5,\ygA) -- (\x + 2.5,\ygB);
	\draw (\x + 3.5,\ygA) -- (\x + 2.5,\ygB);

	\draw (\x + 2.5,\ygA) -- (\x + 3.5,\ygB);
	\draw (\x + 3.5,\ygA) -- (\x + 3.5,\ygB);
	\draw (\x + 4.5,\ygA) -- (\x + 3.5,\ygB);

	\draw (\x + 2.5,\ygB) -- (\x + 2,\ygC);
	\draw (\x + 3.5,\ygB) -- (\x + 2,\ygC);
	\draw (\x + 2.5,\ygB) -- (\x + 3,\ygC);
	\draw (\x + 3.5,\ygB) -- (\x + 3,\ygC);
	\draw (\x + 2.5,\ygB) -- (\x + 4,\ygC);
	\draw (\x + 3.5,\ygB) -- (\x + 4,\ygC);
	
	\draw [dashed] (\x + 0.7, \ygC + 0.8) -- (\x + 5.3, \ygC + 0.8) -- (\x + 5.3, \ygA - 0.8) -- (\x + 0.7, \ygA - 0.8) -- (\x + 0.7, \ygC + 0.8);
	\node at (\x + 4.8,\ygC + 0.1) {$g$};
}

\foreach \x in {-1.5, -0.5, 0.5, 1.5} {
	\draw [->] (\x+3,\yX + 0.65) -- (\x + 3,\ygA - 0.3);
}
\foreach \x in {-1.5, -0.5, 0.5, 1.5} {
	\draw [->] (\x+10,\yX + 0.65) -- (\x + 10,\ygA - 0.3);
}
\foreach \x in {-1.5, -0.5, 0.5, 1.5} {
	\draw [->] (\x+17,\yX + 0.65) -- (\x + 17,\ygA - 0.3);
}

\draw [thick,->,dotted] ($ (w1) - (0,2) $) -- (w1);
\draw [thick,->,dotted] ($ (w2) - (0,2) $) -- (w2);
\draw [thick,->,dotted] ($ (w3) - (0,2) $) -- (w3);

\draw [thick,->,dotted] ($ (b) - (0,2) $) -- (b);
\draw [thick,->,dotted,transform canvas={xshift=-2mm}] ($ (b) - (0,2) $) -- (b);
\draw [thick,->,dotted,transform canvas={xshift=2mm}] ($ (b) - (0,2) $) -- (b);

\draw [thick,->] (3,\ygC) -- (a1);
\draw [thick,->] (10,\ygC) -- (a2);
\draw [thick,->] (17,\ygC) -- (a3);

\draw [thick,->,dotted] (2,\ygC) -- (w1_);
\draw [thick,->,dotted] (9,\ygC) -- (w2_);
\draw [thick,->,dotted] (16,\ygC) -- (w3_);

\draw [thick,->,dotted] (4,\ygC) -- (b1_);
\draw [thick,->,dotted] (11,\ygC) -- (b2_);
\draw [thick,->,dotted] (18,\ygC) -- (b3_);

\end{tikzpicture}

%% file: grbm.bbl
\begin{thebibliography}{10}

\bibitem{Aydin2017}
B.~I. Aydin, Y.~S. Yilmaz, and M.~Demirbas.
\newblock A crowdsourced “who wants to be a millionaire?” player.
\newblock {\em Concurrency and Computation: Practice and Experience}, 2017.

\bibitem{bachrach2012grade}
Y.~Bachrach, T.~Graepel, T.~Minka, and J.~Guiver.
\newblock How to grade a test without knowing the answers --- a bayesian
  graphical model for adaptive crowdsourcing and aptitude testing.
\newblock In {\em 29th Int. Conf. on Machine Learning (ICML)}, pages
  1183--1190, New York, NY, USA, July 2012. Omnipress.

\bibitem{Broelemann2017}
K.~Broelemann, T.~Gottron, and G.~Kasneci.
\newblock Ltd-rbm: Robust and fast latent truth discovery using restricted
  boltzmann machines.
\newblock In {\em IEEE International Conference on Data Engineering}, 2017.

\bibitem{Broelemann2018}
K.~Broelemann, T.~Gottron, and G.~Kasneci.
\newblock Restricted boltzmann machines for robust and fast latent truth
  discovery.
\newblock {\em CoRR}, 2018.

\bibitem{dong2009integrating}
X.~L. Dong, L.~Berti-Equille, and D.~Srivastava.
\newblock Integrating conflicting data: the role of source dependence.
\newblock In {\em VLDB Endowment}, volume~2, pages 550--561. VLDB Endowment,
  2009.

\bibitem{galland2010corroborating}
A.~Galland, S.~Abiteboul, A.~Marian, and P.~Senellart.
\newblock Corroborating information from disagreeing views.
\newblock In {\em 3rd ACM Int. Conf. on Web search and data mining}, pages
  131--140. ACM, 2010.

\bibitem{Garcia-Ulloa2017}
D.~A. Garcia-Ulloa, L.~Xiong, and V.~Sunderam.
\newblock Truth discovery for spatio-temporal events from crowdsourced data.
\newblock {\em Proc. VLDB Endow.}, 10(11):1562--1573, Aug. 2017.

\bibitem{hinton2002training}
G.~E. Hinton.
\newblock Training products of experts by minimizing contrastive divergence.
\newblock {\em Neural Computation}, 14(8):1771--1800, 2002.

\bibitem{hinton2012rbms}
G.~E. Hinton.
\newblock {\em Neural Networks: Tricks of the Trade}, chapter A Practical Guide
  to Training Restricted Boltzmann Machines, pages 599--619.
\newblock Springer, 2012.

\bibitem{Huang2017}
C.~Huang, D.~Wang, and N.~Chawla.
\newblock Scalable uncertainty-aware truth discovery in big data social sensing
  applications for cyber-physical systems.
\newblock {\em IEEE Transactions on Big Data}, 2017.

\bibitem{kasneci2010bayesian}
G.~Kasneci, J.~Van~Gael, R.~Herbrich, and T.~Graepel.
\newblock Bayesian knowledge corroboration with logical rules and user
  feedback.
\newblock In {\em Joint European Conf. on Machine Learning and Knowledge
  Discovery in Databases}, pages 1--18. Springer, 2010.

\bibitem{kasneci2011cobayes}
G.~Kasneci, J.~Van~Gael, D.~Stern, and T.~Graepel.
\newblock Cobayes: bayesian knowledge corroboration with assessors of unknown
  areas of expertise.
\newblock In {\em fourth ACM Int. Conf. on Web search and data mining}, pages
  465--474. ACM, 2011.

\bibitem{li2014confidence}
Q.~Li, Y.~Li, J.~Gao, L.~Su, B.~Zhao, M.~Demirbas, W.~Fan, and J.~Han.
\newblock A confidence-aware approach for truth discovery on long-tail data.
\newblock {\em VLDB Endowment}, 8(4):425--436, 2014.

\bibitem{li2012truth}
X.~Li, X.~Dong, K.~Lyons, W.~Meng, and D.~Srivastava.
\newblock Truth finding on the deep web: Is the problem solved?
\newblock In {\em VLDB Endowment}, volume~6, pages 97--108. VLDB Endowment,
  2012.

\bibitem{li2015survey}
Y.~Li, J.~Gao, C.~Meng, Q.~Li, L.~Su, B.~Zhao, W.~Fan, and J.~Han.
\newblock A survey on truth discovery.
\newblock {\em {SIGKDD} Explorations}, 17(2):1--16, 2015.

\bibitem{Lin2018}
X.~Lin and L.~Chen.
\newblock Domain-aware multi-truth discovery from conflicting sources.
\newblock {\em Proceedings of the VLDB Endowment}, 11(5):635--647, 2018.

\bibitem{wang2013recursive}
D.~Wang, T.~Abdelzaher, L.~Kaplan, and C.~C. Aggarwal.
\newblock Recursive fact-finding: A streaming approach to truth estimation in
  crowdsourcing applications.
\newblock In {\em 33rd Int. Conf. on Distributed Computing Systems (ICDCS)},
  pages 530--539. IEEE, 2013.

\bibitem{wang2014using}
D.~Wang, M.~T. Amin, S.~Li, T.~Abdelzaher, L.~Kaplan, S.~Gu, C.~Pan, H.~Liu,
  C.~C. Aggarwal, R.~Ganti, X.~Wang, P.~Mohapatra, B.~Szymanski, and H.~Le.
\newblock Using humans as sensors: an estimation-theoretic perspective.
\newblock In {\em 13th Int. symposium on Information processing in sensor
  networks}, pages 35--46. IEEE Press, 2014.

\bibitem{wang2012truth}
D.~Wang, L.~Kaplan, H.~Le, and T.~Abdelzaher.
\newblock On truth discovery in social sensing: A maximum likelihood estimation
  approach.
\newblock In {\em 11th Int. Conf. on Information Processing in Sensor
  Networks}, pages 233--244. ACM, 2012.

\bibitem{Yao2018}
L.~Yao, L.~Su, Q.~Li, Y.~Li, F.~Ma, J.~Gao, and A.~Zhang.
\newblock Online truth discovery on time series data.
\newblock 2018.

\bibitem{yin2008truth}
X.~Yin, J.~Han, and S.~Y. Philip.
\newblock Truth discovery with multiple conflicting information providers on
  the web.
\newblock {\em IEEE Trans. Knowl. Data Eng.}, 20(6):796--808, 2008.

\bibitem{Zhang2017}
D.~Y. Zhang, D.~Wang, and Y.~Zhang.
\newblock Constraint-aware dynamic truth discovery in big data social media
  sensing.
\newblock In {\em 2017 IEEE International Conference on Big Data (Big Data)},
  pages 57--66, Dec 2017.

\bibitem{zhao2012probabilistic}
B.~Zhao and J.~Han.
\newblock A probabilistic model for estimating real-valued truth from
  conflicting sources.
\newblock {\em 10th Int. Workshop on Quality in Databases}, 2012.

\bibitem{zhao2012bayesian}
B.~Zhao, B.~Rubinstein, J.~Gemmell, and J.~Han.
\newblock A bayesian approach to discovering truth from conflicting sources for
  data integration.
\newblock {\em VLDB Endowment}, 5(6):550--561, 2012.

\end{thebibliography}
